\documentclass[10pt,twocolumn,letterpaper]{article}
\usepackage{amsmath,amsfonts}
\usepackage{array}
\usepackage{textcomp}
\usepackage{stfloats}
\usepackage{verbatim}
\usepackage{graphicx}
\usepackage{cite}
\usepackage{booktabs}
\usepackage{xcolor}    
\usepackage{colortbl}
\usepackage{arydshln}
\usepackage{tabularx}

\usepackage{multirow}
\usepackage{makecell}
\usepackage{arydshln}
\usepackage{tikz}
\usepackage{multicol}

\makeatletter
\def\blfootnote{\gdef\@thefnmark{}\@footnotetext}
\makeatother
\usepackage{cvpr}              
\definecolor{cvprblue}{rgb}{0.21,0.49,0.74}
\usepackage[pagebackref,breaklinks,colorlinks,allcolors=cvprblue]{hyperref}

\def\paperID{} 
\def\confName{CVPR}
\def\confYear{2026}

\title{Agentic Retoucher for Text-To-Image Generation}

\author{Shaocheng Shen$^{1*}$\footnote[0]{$^{*}$Equal contribution. $^{\dagger}$Corresponding author.}, Jianfeng Liang$^{1*}$, Chunlei Cai$^{1}$, Cong Geng$^{2}$, Huiyu Duan$^{1}$,\\ Xiaoyun Zhang$^{1}$, Qiang Hu$^{1\dagger}$, Guangtao Zhai$^{1}$\\[3pt]
$^{1}$Shanghai Jiao Tong University, Shanghai, China\ $^{2}$JIUTIAN Research, Beijing, China
}

\begin{document}

\twocolumn[{%
\renewcommand\twocolumn[1][]{#1}%
\maketitle

\includegraphics[width=1.0\linewidth]{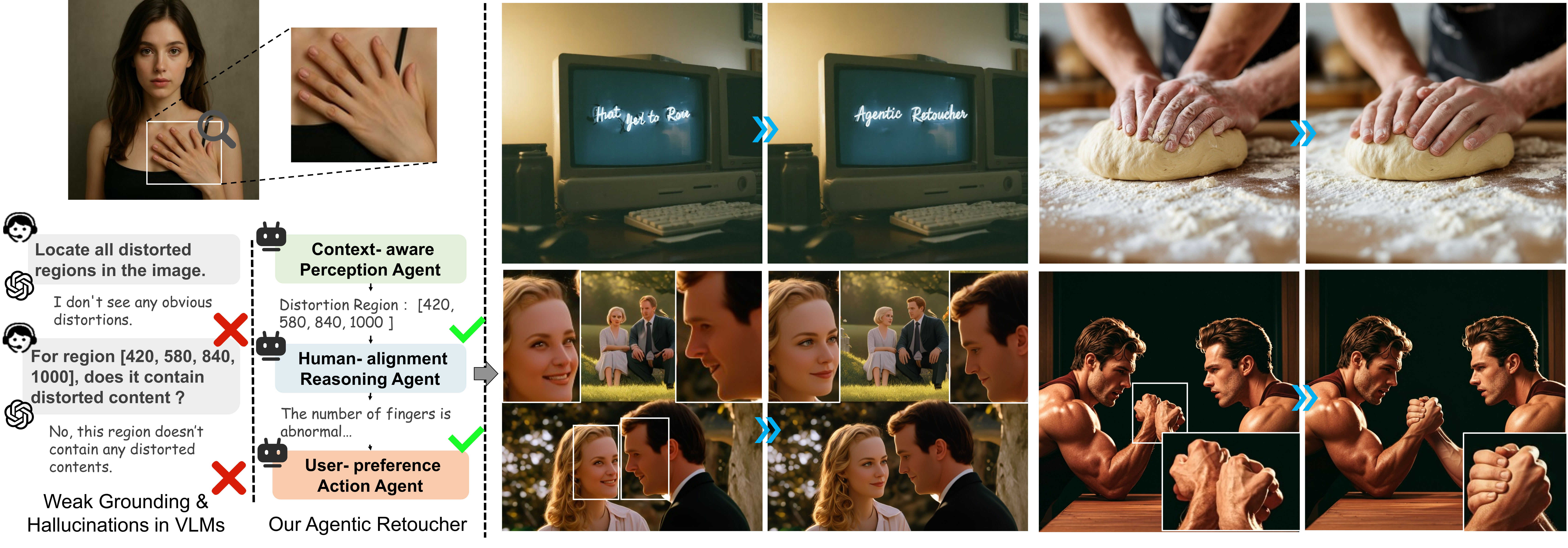}
\captionof{figure}{\textbf{Left:} Existing VLMs hallucinate and fail to localize distortions in AIGC-images, even with explicit region cues, whereas our method accurately localizes distorted regions and provides reasonable diagnoses. \textbf{Right:} Each before-after pair shows the distorted image and the result refined by our Agentic Retoucher, including diverse distortion artifacts across text, hand, face, and interaction. \vspace{1em}}
\label{fig:teaser}
}]

\blfootnote{
 \hspace{-0.575cm} *: These authors contribute equally to this work. \\ 
$\dagger$: Corresponding author.
}

\begin{abstract}

Text-to-image (T2I) diffusion models such as SDXL and FLUX have achieved impressive photorealism, yet small-scale distortions remain pervasive in limbs, face, text and so on. Existing refinement approaches either perform costly iterative re-generation or rely on vision-language models (VLMs) with weak spatial grounding, leading to semantic drift and unreliable local edits. To close this gap, we propose \textbf{Agentic Retoucher}, a hierarchical decision-driven framework that reformulates post-generation correction as a human-like \textit{perception-reasoning-action} loop.
Specifically, we design (1) a \textbf{perception agent} that learns contextual saliency for fine-grained distortion localization under text-image consistency cues, (2) a \textbf{reasoning agent} that performs human-aligned inferential diagnosis via progressive preference alignment, and (3) an \textbf{action agent} that adaptively plans localized inpainting guided by user preference. This design integrates perceptual evidence, linguistic reasoning, and controllable correction into a unified, self-corrective decision process. To enable fine-grained supervision and quantitative evaluation, we further construct \textbf{GenBlemish-27K}, a dataset of 6K T2I images with 27K annotated artifact regions across 12 categories.
Extensive experiments demonstrate that Agentic Retoucher consistently outperforms state-of-the-art methods in perceptual quality, distortion localization and human preference alignment, establishing a new paradigm for self-corrective and perceptually reliable T2I generation.

\end{abstract}    
\section{Introduction}
\label{sec:intro}



Text-to-image (T2I) diffusion models~\cite{fang2025one, liang2026d, fang2025robust} such as Imagen~\cite{saharia2022photorealistic}, Stable Diffusion~\cite{rombach2022high,podell2023sdxl}, FLUX~\cite{flux2024}, and Qwen-Image~\cite{wu2025qwenimagetechnicalreport} have revolutionized image synthesis, enabling photorealistic and creative generation from natural language prompts.
They are now widely adopted in design, film, and entertainment pipelines, as well as in downstream tasks like editing~\cite{brooks2023instructpix2pix,kawar2023imagic,deng2025bagel,liu2025step1x-edit,zhang2025lato} and video generation~\cite{wan2025,opensora,chen2025sanavideoefficientvideogeneration}.
However, even the most advanced models frequently produce small-scale distortions, including misaligned limbs, asymmetric faces, unreadable text, and inconsistent object interactions.
These flaws typically occur locally within otherwise high-quality outputs, making them difficult to detect and expensive to correct through full-image regeneration.
As a result, T2I systems still lack autonomous perceptual reliability, a key barrier to real-world creative and industrial use.



Recent research has explored three main directions to improve generative fidelity: prompt enhancement~\cite{hertz2022prompt,wang2025promptenhancersimpleapproachenhance,wu2025reprompt}, reinforcement learning-based optimization~\cite{black2023ddpo}, and fine-grained noise-space alignment~\cite{izadi2025finegrainedalignmentnoiserefinement,Le2025FromRT,wu2024megafusion}. 
Although these approaches effectively enhance overall realism, they lack explicit spatial reasoning and cannot interpret or correct localized failures. 
Post-hoc editing pipelines such as Imagic~\cite{kawar2023imagic}, Bagel~\cite{deng2025bagel}, and Step1x-Edit~\cite{liu2025step1x-edit} enable local refinement, but rely on manually crafted masks or heuristic textual hints, preventing autonomous identification of regions requiring correction.


Vision-language models (VLMs)~\cite{openai2024gpt4ocard,li2024llavaonevisioneasyvisualtask} show promise as automated critics due to their semantic reasoning capabilities. However, as shown in Fig.~\ref{fig:teaser} (Left), even state-of-the-art VLMs struggle to reliably localize distorted regions. Explicit queries often yield inconsistent or incorrect assessments, with clearly abnormal regions being misjudged as normal. This stems from two key issues: VLMs are optimized for high-level semantic alignment rather than pixel-level verification, leading to weak spatial grounding and missed fine-scale artifacts. Furthermore, their extensive knowledge priors can override visual evidence, causing hallucinated judgments. For example, a portrait with six fingers is deemed plausible despite explicit highlighting of the defective hand, demonstrating that current VLMs are unreliable for fine-grained artifact detection in AI-generated images.

To address these limitations, we present \textbf{Agentic Retoucher}, a hierarchical decision-driven framework that reformulates post-generation correction as a structured \textit{perception-reasoning-action} loop. Agentic Retoucher comprises three collaborative agents that execute a unified self-refinement cycle. The perception agent predicts context-aware distortion saliency by integrating visual evidence with prompt semantics, generating reliable region proposals for fine-scale anomalies. The reasoning agent performs human-aligned diagnostic inference, including identifying distortion types, detailing the appearance of distortions and assessing their inconsistency with global images through progressive preference alignment. The action agent then adaptively selects and executes targeted retouching operations from a modular tool library, supporting both mask-guided and instruction-driven editing under user or environment constraints. Through iterative verification, these components fuse perceptual cues, semantic reasoning, and controllable tool-based correction into a coherent self-corrective process, enabling the proposed Agentic Retoucher automatically refine distortion artifacts across text, hand, face, and interaction. (Fig.~\ref{fig:teaser}, Right)

To enable fine-grained and region-aware supervision, we construct \textbf{GenBlemish-27K}, a dataset of 6K T2I images with 27K pixel-level annotated distortion regions spanning 12 representative artifact categories. This dataset provides both spatial grounding and semantic diagnostic cues, allowing our system to reliably map localized distortions to interpretable region-level feedback and convert them into targeted retouching actions. Beyond supporting our framework, GenBlemish-27K also improves VLM robustness for evaluating AIGC imagery, enhancing region-grounded assessment and steering adaptation toward human-aligned distortion reasoning. Extensive experiments show that Agentic Retoucher significantly boosts local perceptual fidelity across diverse diffusion backbones while preserving global coherence, outperforming state-of-the-art post-editing methods on both objective metrics (plausibility score increasing from 44.21 to \textbf{47.10}) and human preference studies (\textbf{83.2\%} preferred over unretouched outputs).


Our main contributions are summarized as follows:
\begin{itemize}
    \item We propose \textbf{Agentic Retoucher}, a novel paradigm that reformulates post-generation editing as a \textit{perception-reasoning-action} loop, enabling diffusion models to autonomously diagnose and refine their artifacts. 
    \item  We design a collaborative three-agent system, where a perception agent performs context-aware distortion localization, a reasoning agent conducts human-aligned fine-grained diagnosis, and an action agent performs adaptive local retouching with user-guided tools.
    \item We construct the \textbf{GenBlemish-27K} with pixel-level masks and textual annotations across 12 artifact types, providing a dataset for fine-grained artifact perception and correction. 
    \item Extensive experiments demonstrate that our framework achieves state-of-the-art performance in perceptual quality enhancement, artifact localization and textual description accuracy across diverse diffusion backbones. 
\end{itemize}

~\section{Related Works}
\label{RW}

\begin{figure*}[t]
  \centering
  \includegraphics[width=1.0\linewidth]{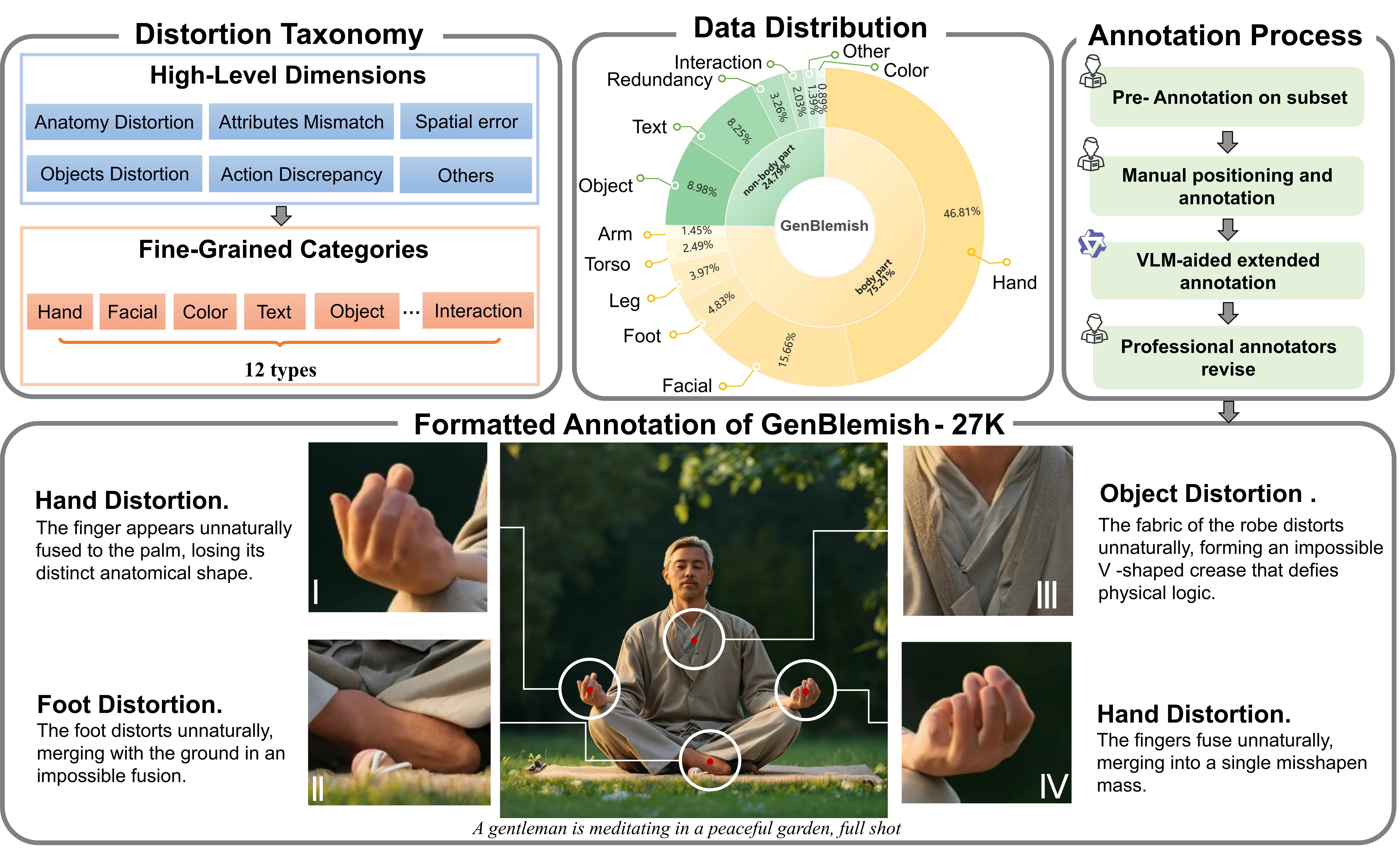}
  \caption{Overview of GenBlemish-27K. The figure illustrates (a) the dual-layer distortion taxonomy with six high-level dimensions and twelve fine-grained categories, (b) the distribution of localized distortion types, (c) the human-AI collaborative annotation pipeline, and (d) representative formatted samples with pixel-level masks and textual descriptions, highlighting how GenBlemish-27K enables fine-grained localization and reasoning over diverse text-to-image distortions.}
  \label{fig:dataset}
  \vspace{-10pt}
\end{figure*}

\textbf{Visual Quality Assessment.} Visual Quality Assessment (VQA)~\cite{qian2025explainablepartialaigcimagequality,Liang_2024_CVPR,cao2024synartifact,chang2025oneigbenchomnidimensionalnuancedevaluation,duan2025finevq,liu2025f} is an important and rapidly evolving field that has made significant contributions to evaluating a wide range of image and video tasks with closer alignment to human subjective perception. For AIGC content assessment, most existing work~\cite{wang2025cigeval,zhang2024abench} is limited to applying quantitative metrics at global scales, without explicit localization and assessment of local flaws. RichHF~\cite{Liang_2024_CVPR} introduces predictors for local structural distortions along with a corresponding scoring procedure. However, these methods focus solely on assessment and have not been integrated into an automated, closed-loop pipeline for evaluation and refinement.


\textbf{Vision-Language Model (VLMs).} VLMs~\cite{vteam2025glm45vglm41vthinkingversatilemultimodal,yang2025qwen3technicalreport,lu2025ovis25technicalreport} have become leading drivers of general artificial intelligence, exhibiting remarkable problem-solving and reasoning ability through training on large-scale multimodal data (e.g., GPT-4o~\cite{GPT4} and the Qwen~\cite{yang2025qwen3technicalreport} family in real-world multimodal interaction). Several works~\cite{kaduri2024_vision_of_vlms, Qu_2025_CVPR,Heo_2025_CVPR} further advance VLMs on image understanding tasks. However, heavy reliance on high-fidelity pretraining data and learned priors often biases VLMs toward prior-based, ungrounded hallucinated responses in the context of text-to-image evaluation. 


\textbf{Agentic System in Vision.} Agentic System~\cite{538972,RotherKB04,Cubuk2019AutoAugmentLA,wang2021tent,zhu2023ghost,li2023generalist,chen2024restoreagent} adopts active, closed-loop perception-decision-action framework, with VLMs increasingly acting as planners due to their strong reasoning. In the 3D domain, VADAR~\cite{Marsili_2025_CVPR} proposes an agentic program synthesis approach, achieving superior performance in 3D spatial reasoning. In image and video restoration, AgenticIR~\cite{agenticir} and MoA-VR~\cite{liu2025moavrmixtureofagentsallinonevideo} independently propose VLM-integrated multi-agent repair paradigms. In the realm of artistic creation, JarvisArt~\cite{jarvisart2025} enables fine-grained photo retouching via tool invocation based on user instructions.

\section{Dataset: GenBlemish-27K}
\begin{figure*}[t]  
    \centering
    \includegraphics[width=1.0\textwidth]{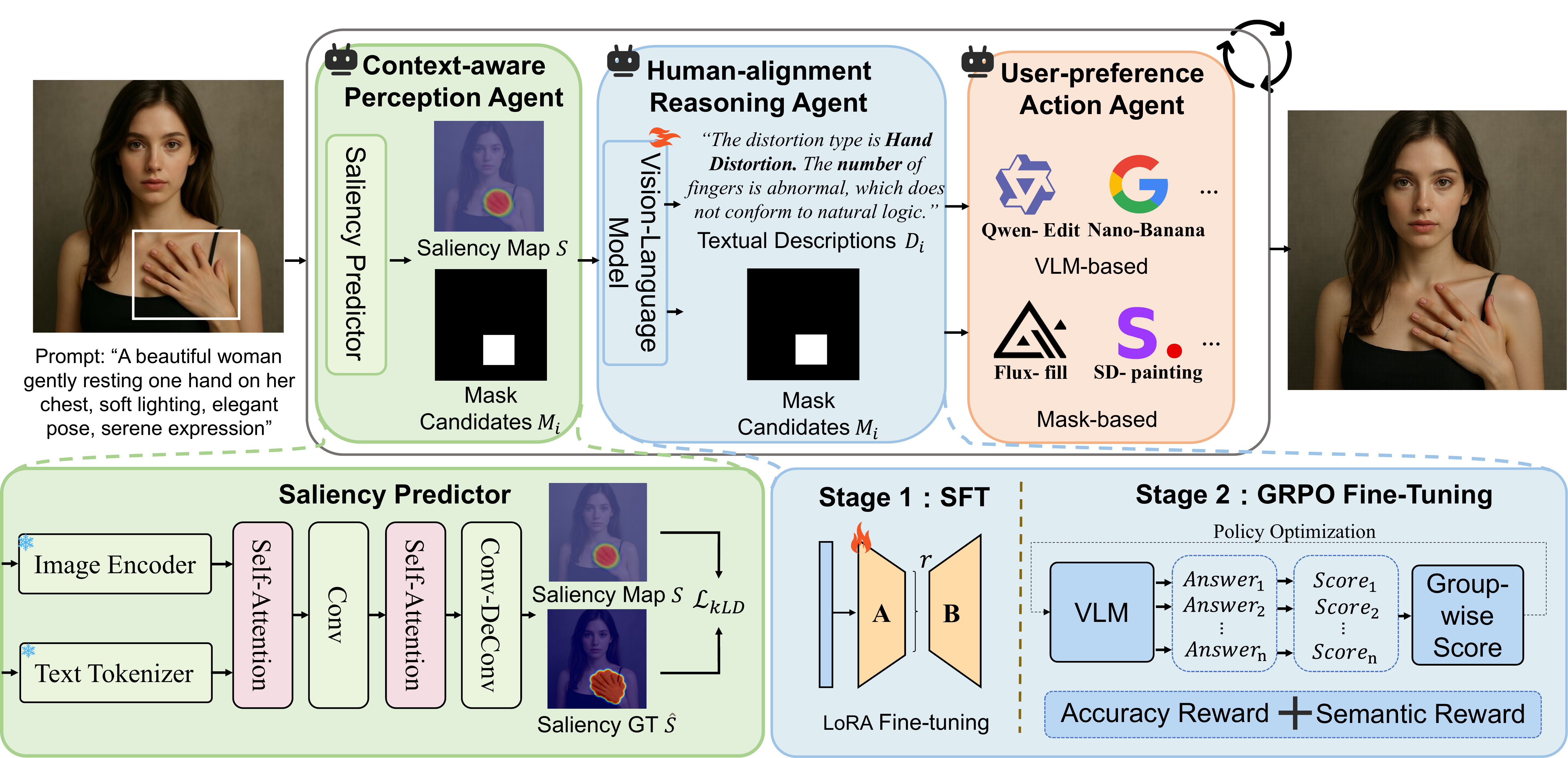}  
    \caption{Overview of the proposed Agentic Retoucher. The framework operates as a \textit{perception-reasoning-action} loop for post-generation correction in AIGC. The Perception Agent localizes context-dependent distortions via cross-modal saliency prediction, the Reasoning Agent performs human-aligned diagnosis through iterative reasoning, and the Action Agent executes adaptive localized inpainting guided by reasoning outputs, forming a closed-loop self-corrective process.}  
    \label{Fig:pipeline}
    \vspace{-10pt}
\end{figure*}

We construct GenBlemish-27K, a large-scale dataset designed for granular distortion diagnosis and reasoning in text-to-image generation. 
It provides pixel-level annotations and natural-language descriptions for over 27K distorted regions across 12 artifact types, offering comprehensive supervision for perception, reasoning, and localized correction tasks.

\subsection{Distortion Taxonomy}
Existing T2I evaluation datasets suffer from limited coverage (e.g., HADM~\cite{Wang2024HADM}, Wang et al.~\cite{Wang_2025_CVPR}), coarse annotation (e.g., RichHF~\cite{Liang_2024_CVPR}), and insufficient scale (e.g., SynArtifacts-1K~\cite{cao2024synartifact} with only 1K samples). 
To address these issues, GenBlemish-27K establishes a hierarchical taxonomy of distortions derived from large-scale inspection of outputs from mainstream T2I models. 
We define six high-level distortion dimensions including human anatomical distortion, attribute inconsistency, spatial errors, object deformation or redundancy, action and interaction distortion, and miscellaneous cases. 
These dimensions are further refined into 12 fine-grained categories such as limb deformities, face distortion, and text anomalies. 
This taxonomy captures typical artifacts observed in state-of-the-art diffusion models and enables interpretable reasoning about both \textit{what} and \textit{where} fails (see Fig.~\ref{fig:dataset}).

\subsection{Data Annotation}
We curate 6,025 images from EvalMuse-Structure~\cite{han2025ntire2025challengetext}, covering outputs from over 20 T2I models such as Dreamina, Midjourney, Kandinsky~\cite{arkhipkin2023kandinsky}, and SDXL~\cite{podell2023sdxl}. 
A four-stage human-in-the-loop process ensures both semantic richness and annotation consistency. 
(1) Annotators are first calibrated through a pre-annotation stage. 
(2) For each distortion region, multiple annotators independently provide the center, category, and a brief textual description; the region radius is 1/20 of the image height. 
(3) The textual descriptions are expanded and refined using QwenVL-Max~\cite{Qwen2.5-VL}. 
(4) Final annotations are reconciled through majority voting and expert validation. 
Each sample includes the generated image, input prompt, distortion mask, category label, and natural-language description, supporting tasks such as saliency prediction, defect classification, and language grounding.

\subsection{Dataset Statistics}
GenBlemish-27K consists of 6,025 images, 27,507 annotated distortion regions. The agreement rate between majority voting and expert validation exceeds 95\%, confirming annotation reliability. 
Each image contains an average of 4.6 annotated regions, each paired with an 11.8-word description. 
As shown in Fig.~\ref{fig:dataset}, hand distortions account for 46.8\% of all annotations, followed by facial defects at 15.7\%. 
These statistics indicate that fine-grained human generation remains a persistent challenge even for advanced diffusion models. More details will be included in supplementary material.

\section{Methodology}
\label{sec:method}

\subsection{Overview of the Agentic Retoucher}

We propose an \textbf{Agentic Retoucher} that redefines post-generation image correction as a closed \textit{perception-reasoning-action} loop.
Unlike conventional feed-forward editing pipelines that apply static refinement, our framework introduces autonomy, interpretability, and self-correction into the editing process.
By framing retouching as a sequential decision process, the model can reason about \textit{what} and \textit{where} distortions occur before performing targeted correction, bridging perceptual evidence, semantic inference and controllable within a unified architecture.

As shown in Fig.~\ref{Fig:pipeline}, the framework consists of three collaborative agents.
The Perception Agent detects context-dependent distortions from visual-textual cues and generates a distortion-saliency map.
The Reasoning Agent analyzes the detected regions, identifies distortion categories, and produces human-aligned textual descriptions.
The Action Agent executes localized retouch guided by reasoning outputs, closing the \textit{perception-reasoning-action} cycle through iterative refinement.

Formally, let $I_t$ denote the image to be retouched. At iteration $t$, the perception agent produces a saliency map $S_t$ highlighting anomalous regions.
If the saliency $S_t$ exceeds a threshold $\tau_s$, the reasoning agent infers distortion types and generates region-level descriptions $\{D_i\}$ and masks $\{M_i\}$.
The action agent then applies localized refinement to obtain an updated image:
\begin{equation}
    I_{t+1} = \Phi_{\text{act}}(I_t, \{M_i \lor D_i\}), \quad t \leftarrow t + 1.
\end{equation}
This process repeats until all salient distortions are eliminated, producing a perceptually faithful result.
Through this iterative loop, the framework transitions post-generation editing from reactive correction to proactive reasoning, integrating perceptual analysis, contextual understanding, and controllable retouching in a single interpretable pipeline.


\subsection{Context-Aware Perceptual Distortion Analysis}
\label{sec:perception}

Text-to-image generations frequently exhibit subtle and context-dependent distortions such as implausible limb, object and text.
These artifacts often lack explicit object boundaries, making conventional pixel-wise detection unreliable.
To emulate human perceptual sensitivity, inspired by~\cite{Liang_2024_CVPR}, we design a context-aware saliency predictor that estimates a distortion-saliency map $S \in [0,1]^{H\times W}$ conditioned on both the generated image $I$ and its prompt $P$.
A dual-encoder ViT~\cite{2020An}-T5~\cite{roberts2022scalingmodelsdatatextttt5x} backbone encodes image and text representations, which are subsequently concatenated and fused via a self-attention mechanism to capture inherent correspondences between visual structures and textual semantics.
A lightweight attention refinement module further aggregates multi-scale contextual cues, improving the detection of distortions whose visibility depends on global images.

The model is optimized using a hybrid loss that balances pixel accuracy and distributional consistency:
\begin{equation}
    \mathcal{L}_{\text{sal}} = \alpha \mathcal{L}_{\text{MSE}}(S, \hat{S}) + (1-\alpha)\mathcal{L}_{\text{KLD}}(S, \hat{S}),
\end{equation}
where $\hat{S}$ is the ground-truth saliency and $\alpha$ controls the balance between reconstruction precision and perceptual alignment.
The KLD term encourages alignment with human fixation distributions, preserving discriminability in ambiguous regions and preventing over-smoothing.
The resulting saliency map is binarized and morphologically dilated to form mask candidates $\{M_i\}$ for subsequent reasoning.

Beyond localization, the predicted saliency reflects contextual anomalies, serving as an explicit spatial prior that helps the reasoning agent focus on regions requiring further analysis, ensuring that higher-level diagnosis emerges from low-level visual awareness.

\subsection{Human-Aligned Reasoning and Adaptive Action}
\label{sec:reasoning}

Given localized regions $\{M_i\}$, the reasoning agent performs inferential diagnosis to generate textual descriptions $\{D_i\}$ that capture distortion types, local characteristics and contextual relationships.
This task requires structured reasoning aligned with human perceptual judgment rather than simple classification or captioning.
We adopt a progressive preference alignment paradigm consisting of two complementary stages: supervised fine-tuning (SFT) for structural initialization and Group Relative Policy Optimization (GRPO) for human-aligned reinforcement.

In the first stage, SFT establishes structured response formats and distortion taxonomy under limited supervision.
To reduce computational overhead, we employ Low-Rank Adaptation (LoRA)~\cite{hu2022lowrank}, where the weight update $\Delta W$ for a layer $W \in \mathbb{R}^{n\times m}$ is decomposed as $\Delta W = AB$, with $A \in \mathbb{R}^{n\times r}$ and $B \in \mathbb{R}^{r\times m}$.
This low-rank decomposition enables efficient specialization of the reasoning model without full-parameter fine-tuning.

In the second stage, GRPO~\cite{deepseekai2025deepseekr1incentivizingreasoningcapability} aligns reasoning behavior with human preferences through reinforcement signals:

\begin{align}
    \mathcal{L}_{\text{GRPO}} =\ &\mathbb{E}_{(q,o)}[\min(r_t \hat{A}_t,\, \mathrm{clip}(r_t, 1-\varepsilon, 1+\varepsilon)\hat{A}_t) \notag\\
    &\qquad\quad - \beta D_{\text{KL}}[\pi_\theta || \pi_{\text{ref}}] ],
\end{align}
where $\hat{A}_t$ denotes the normalized advantage that captures preference consistency. Policy optimization is guided by rewards capturing distortion-type classification accuracy and alignment between textual descriptions and human labels.
This stage reduces hallucination, enabling the agent to produce consistent, human-aligned reasoning across diverse distortion patterns.

Building on the reasoning outputs, the Action Agent transforms $\{M_i, D_i\}$ into controllable local editing operations.
It determines the spatial extent, tool selection, and inpainting instruction for each region.
Depending on computational constraints or user preferences, the agent dynamically chooses between VLM-based or mask-guided inpainting from a modular tool library.
The updated image is re-evaluated by the perception agent to
close the \textit{perception-reasoning-action} loop.
Through iterative perception and reasoning, the framework converges toward high-quality outputs with plausible details.

The proposed framework transitions post-generation editing from reactive correction to proactive reasoning.
By integrating perception-driven diagnosis, human-aligned reasoning, and adaptive retouching into a unified loop, the system achieves interpretable and autonomous refinement of generative outputs.

\section{Experiments}

\begin{table*}[t]
\centering
\caption{Quantitative comparison of Agentic Retoucher with VLM-based and mask-based inpainting baselines on the GenBlemish-27K and SynArtifacts-1K datasets.} 
\renewcommand\arraystretch{1.2}
\resizebox{1.0\textwidth}{!}{
\begin{tabular}{cccccc|cccc}
\toprule[1pt]
\multicolumn{1}{l}{\textbf{Condition Type}} & \textbf{Model}                         & \multicolumn{4}{c}{\textbf{GenBlemish-27K}}                                                                                                                                                                     & \multicolumn{4}{c}{\textbf{SynArtifacts-1K}}                                                                                                                                                                   \\ \hline
\multicolumn{1}{l}{}                        & \multicolumn{1}{l}{}                   & \multicolumn{1}{l}{\textbf{plausibility$\uparrow$}} & \multicolumn{1}{l}{\textbf{aesthetics$\uparrow$}} & \multicolumn{1}{l}{\textbf{alignment$\uparrow$}} & \multicolumn{1}{l}{\textbf{overall$\uparrow$}} & \multicolumn{1}{l}{\textbf{plausibility$\uparrow$}} & \multicolumn{1}{l}{\textbf{aesthetics$\uparrow$}} & \multicolumn{1}{l}{\textbf{alignment$\uparrow$}} & \multicolumn{1}{l}{\textbf{overall$\uparrow$}} \\ \cline{3-10} 
\multicolumn{1}{l}{}                        & Original                               & 44.21                                               & 53.69                                             & 57.89                                            & 47.15                                          & 61.53                                               & 61.63                                             & 60.65                                            & 55.35                                          \\ \cdashline{2-10}
\multirow{4}{*}{\textbf{VLM-based}}         & Qwen-Edit                              & 44.44                                               & 53.71                                             & 57.69                                            & 47.15                                          & 61.45                                               & 61.64                                             & 60.70                                            & 55.33                                          \\
                                            & \cellcolor{gray!20}\textbf{Ours w Qwen-Edit}              & \cellcolor{gray!20}\textbf{47.10}                                      & \cellcolor{gray!20}\textbf{55.75}                                    & \cellcolor{gray!20}\textbf{59.54}                                   & \cellcolor{gray!20}\textbf{49.27}                                 & \cellcolor{gray!20}\textbf{65.43}                                                     & \cellcolor{gray!20}\textbf{64.88}                                                    &  \cellcolor{gray!20}\textbf{62.61}                                                  &  \cellcolor{gray!20}\textbf{58.04}                                                \\\cdashline{2-10}
                                            & Gemini 2.5 Flash Image                 & 44.41                                               & 53.80                                             & 57.93                                            & 47.27                                          & 62.63                                               & 63.07                                             & 61.21                                            & 56.15                                          \\
                                            & \cellcolor{gray!20}\textbf{Ours w Gemini 2.5 Flash Image} & \cellcolor{gray!20} \textbf{46.81}                                      & \cellcolor{gray!20}\textbf{55.47}                                    & \cellcolor{gray!20}\textbf{59.22}                                   & \cellcolor{gray!20}\textbf{48.97}                                 & \cellcolor{gray!20}\textbf{65.96}                                & \cellcolor{gray!20}\textbf{65.27}                             & \cellcolor{gray!20}\textbf{62.94}                            & \cellcolor{gray!20}\textbf{58.43}                         \\ \hline
\multirow{4}{*}{\textbf{Mask-based}}        & Flux-fill                              & 44.12                                               & 53.68                                             & 57.91                                            & 47.07                                          & 61.92                                               & 61.78                                             & 61.17                                            & 55.71                                          \\
                                            & \cellcolor{gray!20}\textbf{Ours w Flux-fill}              & \cellcolor{gray!20}\textbf{46.18}                                      & \cellcolor{gray!20}\textbf{55.17}                                    & \cellcolor{gray!20}\textbf{59.26}                                   & \cellcolor{gray!20}\textbf{48.66}                                 & \cellcolor{gray!20}\textbf{65.25}                                      & \cellcolor{gray!20}\textbf{64.07}                                    & \cellcolor{gray!20}\textbf{62.74}                                   & \cellcolor{gray!20}\textbf{57.86}                                 \\\cdashline{2-10}
                                            & SD-inpainting                          & 45.18                                               & 53.85                                             & 57.70                                            & 47.50                                          & 63.60                                               & 62.49                                             & 60.88                                            & 56.26                                          \\
                                            & \cellcolor{gray!20}\textbf{Ours w SD-inpainting}          & \cellcolor{gray!20}\textbf{46.71}                                      & \cellcolor{gray!20}\textbf{54.71}                                    & \cellcolor{gray!20}\textbf{58.07}                                   & \cellcolor{gray!20}\textbf{48.31}                                 & \cellcolor{gray!20}\textbf{66.66}                                      & \cellcolor{gray!20}\textbf{64.67}                                    & \cellcolor{gray!20}\textbf{62.33}                                   & \cellcolor{gray!20}\textbf{58.27}                                 \\ \bottomrule[1pt]
\end{tabular}
}
\label{tab:table1}
\vspace{-10pt}
\end{table*}

\begin{table}[t]
  \centering
  \normalsize 
  \caption{Human evaluation results: preference distribution comparing Agentic Retoucher outputs to original images. Percentages of test cases rated as $\gg$ (significantly better), $>$ (slightly better), $\approx$ (about the same), $<$ (slightly worse), or $\ll$ (significantly worse). Data from 5 participants in a randomized, blind survey.}

  \begin{tabular*}{0.45\textwidth}{@{\extracolsep{\fill}}lccccc} 
    \toprule[1pt]
    Preference & $\gg$ & $>$ & $\approx$ & $<$ & $\ll$ \\
    \midrule
     Baseline & 4.2\% & 22.8\% & 60.8\% & 9.2\% & 3.0\% \\
     \textbf{Ours} & 48.8\% & 34.4\% & 10.2\% & 5.8\% & 0.8\% \\
    \bottomrule[1pt]
  \end{tabular*}
  \label{tab:HumanPreference}
  \vspace{-11pt}
\end{table}

\subsection{Experimental setup}

\textbf{Datasets and Implementation Details.}
We evaluate our framework on the proposed GenBlemish-27K dataset to assess optimization efficacy and module functionality, and further verify its generalization on SynArtifacts-1K~\cite{cao2024synartifact}.
For the Context-Aware Perception Agent, the learning rate is set to $2\times10^{-5}$.
For the Human-Alignment Reasoning Agent, Stage 1 employs LoRA fine-tuning with rank 64 and $\alpha=32$.
Inference is fully automatic and converges within 2-3 reasoning iterations per image.


\textbf{Evaluation Metrics.}
For retouch evaluation, we use the four perceptual metrics from RichHF~\cite{Liang_2024_CVPR}: plausibility, aesthetics, alignment, and overall, to assess both structural plausibility and perceptual quality, as standard T2I metrics (e.g., FID~\cite{NIPS2017_8a1d6947}) fail to capture localized improvements.
For perception, following \cite{8315047}, we adopt CC, SIM, KLD, AUC-Judd, and NSS to evaluate distributional and fixation-level consistency.
For reasoning, we report distortion-type classification accuracy and semantic alignment using ROUGE~\cite{lin-2004-rouge}, METEOR~\cite{banerjee-lavie-2005-meteor}, Word2Vec~\cite{mikolov2013efficientestimationwordrepresentations}, and SimCSE~\cite{gao2021simcse}.
More setup details are provided in the supplementary material.

\subsection{Comparison}

\textbf{Quantitative comparisons.} We evaluate Agentic Retoucher using two categories of inpainting tools within our Adaptive Action Toolkit: VLM-based models (Qwen-Edit and Gemini 2.5 Flash Image) and mask-based models (Flux-Fill and SD-inpainting). As shown in Tab.~\ref{tab:table1}, Agentic Retoucher consistently surpasses all baselines across the four perceptual metrics of plausibility, aesthetics, alignment, and overall.
On the GenBlemish-27K, the plausibility score improves from 44.21 to \textbf{47.10} and the overall score from 47.15 to \textbf{49.27}, indicating that our system effectively handles localized distortion while preserving global structure and style. Similar improvements are observed on SynArtifacts-1K (overall score reaching \textbf{58.43}), confirming the generalization capability of the proposed framework. In human evaluations, as shown in Tab.~\ref{tab:HumanPreference}, the Agentic Retoucher substantially outperforms the baseline model, with \textbf{83.2\%} of results judged superior to the pre-retouching images, further demonstrating the visual expressiveness of our method.
\begin{figure*}[t]  
    \centering
    \setlength{\abovecaptionskip}{6pt}
    \includegraphics[width=1.0\textwidth]{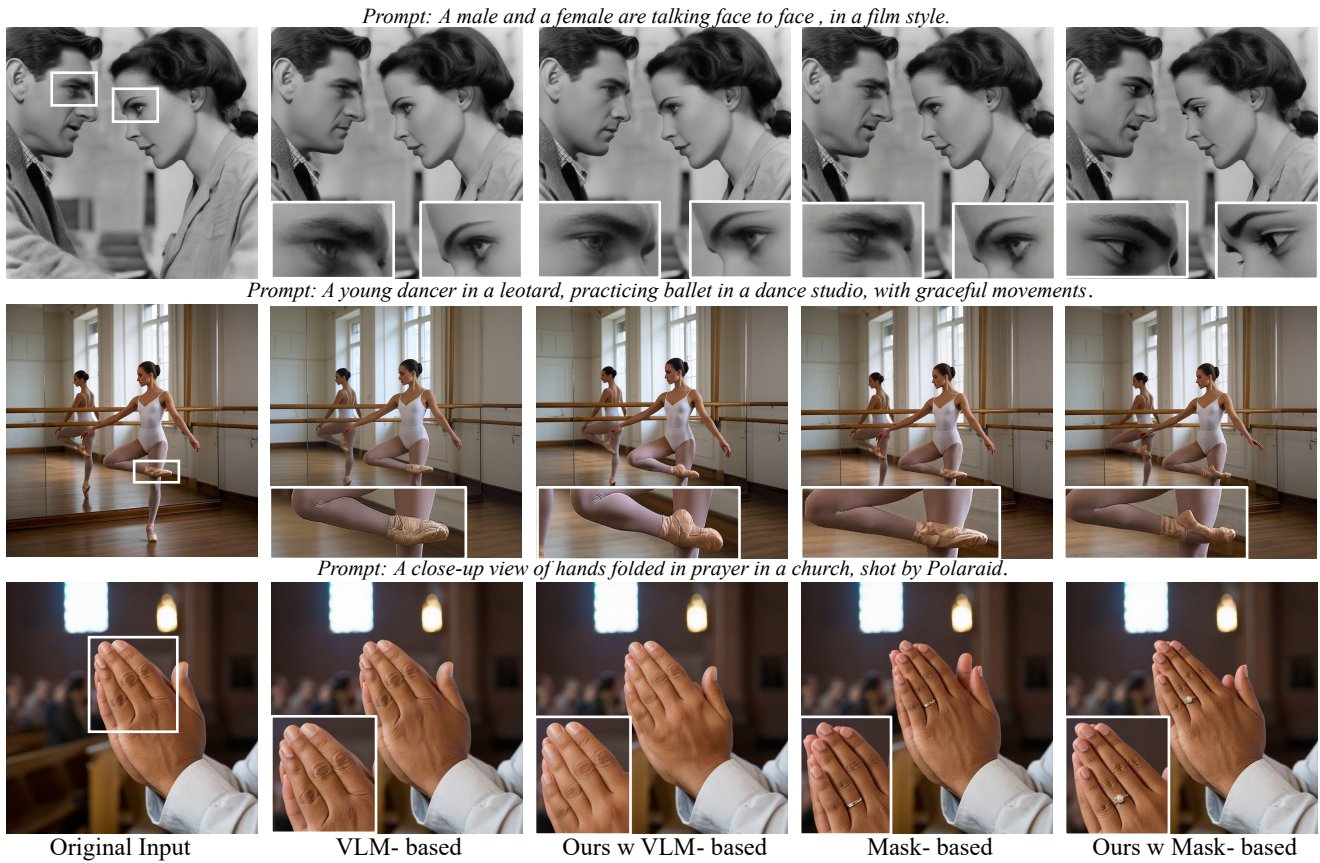}  
    \caption{Qualitative comparison of retouching results across diverse prompts.
White bounding boxes indicate zoomed-in fine-grained regions.
Agentic Retoucher retouches local distortions and implausibilities while maintaining global visual harmony, outperforming both VLM-based and mask-based baselines. More qualitative results are included in the supplementary material.}
    \label{Fig:results}
    \vspace{-10pt}
\end{figure*}

\begin{table}[t]
  \centering
  \caption{Quantitative evaluation of the Context-Aware Perception Agent on distortion-aware saliency prediction.
Higher AUC-Judd, NSS, CC, SIM and lower KLD indicate better context perception.}
  \renewcommand\arraystretch{1.2}  
  \resizebox{0.5\textwidth}{!}{  
    \begin{tabular}{lcccccc}  
      \toprule[1pt]
      \textbf{Method/Metric} & \textbf{AUC-Judd} $\uparrow$ & \textbf{ NSS}$\uparrow$ & \textbf{ CC} $\uparrow$& \textbf{ SIM}$\uparrow$ &\textbf{ KLD}$\downarrow$ \\
      \midrule
      AIM~\cite{NIPS2005_0738069b} & 0.7822 & 1.1479 & 0.1667 & 0.0759 & 3.0185 \\
      GBVS~\cite{NIPS2006_4db0f8b0} & 0.6580 & 0.5811 & 0.0080 & 0.0010 & 8.5547 \\
      SR~\cite{4270292} & 0.5336 & 0.0135 & 0.0002 & 0.0524 & 3.4162 \\
      SMVJ~\cite{NIPS2007_708f3cf8} & 0.8167 & 0.7121 & 0.0778 & 0.0623 & 3.3722 \\
      SWD~\cite{zhao2011visual} & 0.8170 & 0.5307 & 0.0712 & 0.0761 & 3.5233 \\
      CA~\cite{5539929} & 0.4516 & 0.7633 & 0.1000 & 0.0693 & 3.3286 \\
      \hdashline
      SALICON~\cite{Huang_2015_ICCV} & 0.9230 & 1.0774 & 0.5039 & 0.2734 & 1.7171\\
      TranSalNet~\cite{Lou2022TranSalNetTP} & 0.9029 & 1.1494 & 0.4616 & 0.0989  & 2.8716\\
      Sal-CFS-GAN~\cite{Che_2020} & 0.7747 & 0.7810 & 0.2124 & 0.1018 & 2.8589 \\
      SAM-VGG~\cite{cornia2018predicting} & 0.8773 & \underline{1.2072} & 0.3410 & 0.1791 & 2.4094 \\
      SAM-ResNet~\cite{cornia2018predicting} & 0.9162 & 1.0552 & 0.4040 & 0.2475 & 2.0740 \\
      MLNet~\cite{mlnet2016} & 0.8539 & 1.0455 & 0.3535 & 0.2381 & 2.2359 \\
      \hdashline
      InternVL3.5-8B~\cite{wang2025internvl35advancingopensourcemultimodal}  & 0.8049 & 0.7689 & \underline{0.5104} & \textbf{0.4095} & 3.9325 \\
      Qwen2.5-VL-7B~\cite{Qwen2.5-VL} & 0.6145 & 0.4190 & 0.1710 & 0.1481 & 7.4353 \\
      GLM-4.1V-9B~\cite{vteam2025glm45vglm41vthinkingversatilemultimodal} & 0.5461 & 0.2191 & 0.0604 & 0.0902 & 8.0118\\
      \hdashline
      RichHF~\cite{Liang_2024_CVPR} & \underline{0.9211} & 0.8954 & 0.4748 & 0.3309 & \underline{1.6697} \\
      \rowcolor{gray!20} 
      Ours & \textbf{0.9336} & \textbf{1.2087} & \textbf{0.5568} & \underline{0.3822} & \textbf{1.4313} \\
      \bottomrule[1pt]
    \end{tabular}
  }
  \label{tab:Saliency map}
  \vspace{-15pt}
\end{table}

\textbf{Qualitative comparisons.} Fig.~\ref{Fig:results} presents qualitative comparisons across various scenes and prompt conditions.
Agentic Retoucher autonomously identifies distortion regions and performs targeted refinement while preserving global composition.
Zoomed-in crops reveal that our method excels at retouching fine-grained geometric details (e.g.,faces, fingers, feet), with coherent shading and natural boundaries.
In contrast, VLM-based methods fail to localize distortions, and their retouch performance degrades without fine-grained instruction guidance, whereas mask-based models revert to stochastic refinement once explicit masks are removed.
These results highlight the effectiveness of our agentic perception-reasoning-action loop in achieving both localized precision and holistic consistency.

\begin{figure}[t]  
    \centering
    \includegraphics[width=0.5\textwidth]{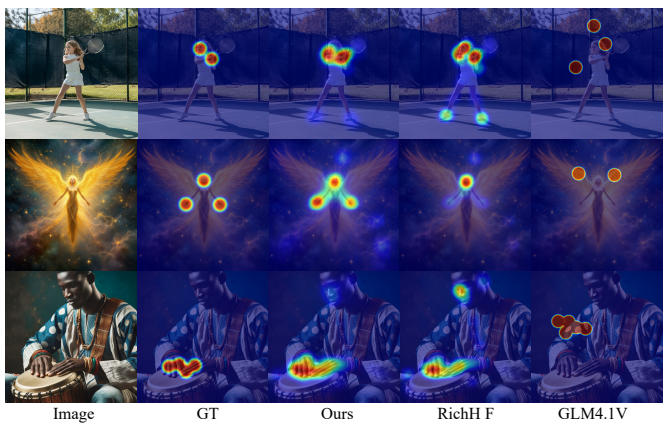}  
    \caption{Qualitative visualization of saliency prediction. Our method yields sharper, context-aware localization than RichHF and GLM4.1V.}
    \label{Fig:Saliency map}
    \vspace{-20pt}
\end{figure}

\subsection{Perception and Reasoning Analysis}

\textbf{Context-Aware Perception.}  Tab.~\ref{tab:Saliency map} compares the proposed Context-Aware Perception Agent with conventional saliency detectors, deep saliency networks, and vision-language models. Hand-crafted methods (e.g., AIM~\cite{NIPS2005_0738069b}, GBVS~\cite{NIPS2006_4db0f8b0}) rely on low-level contrast, while deep saliency networks (e.g., SAM-ResNet~\cite{cornia2018predicting}, TranSalNet~\cite{Lou2022TranSalNetTP}) yield only moderate gains; both fail to capture context-aware distortion regions.
General-purpose VLMs (e.g., InternVL3.5~\cite{wang2025internvl35advancingopensourcemultimodal}, Qwen2.5-VL~\cite{Qwen2.5-VL}, GLM-4.1V~\cite{vteam2025glm45vglm41vthinkingversatilemultimodal}) also perform poorly, lacking task-specific grounding for distortion awareness.
In contrast, our perception agent achieves state-of-the-art results across all metrics (AUC-Judd = 0.9336, NSS = 1.2087, KLD = 1.4313), demonstrating robust localization of artifact-prone regions.


Fig.~\ref{Fig:Saliency map} shows that, unlike RichHF (overemphasizing facial and limb regions) and GLM4.1V (dispersing attention to irrelevant areas), our saliency maps deliver sharper spatial focus and stronger agreement with ground-truth labels.This precise localization provides a reliable foundation for subsequent reasoning and retouching within the loop.

\begin{table}[t]
  \centering
  \caption{Quantitative evaluation and ablation of the Human-Alignment Reasoning Agent}
  \renewcommand\arraystretch{1.3}  
  \resizebox{0.5\textwidth}{!}{  
    \setlength{\tabcolsep}{4pt} 
    \begin{tabular}{lcccccccc}  
      \toprule[1pt]
      \normalsize
      \textbf{Method/Metric} & \textbf{Accuracy} $\uparrow$ & \textbf{ SimCSE}$\uparrow$ & \textbf{Word2Vec}$\uparrow$ & \textbf{Meteor} $\uparrow$ & \textbf{ROUGE} $\uparrow$ \\
      \midrule
      GPT 5 Zero-Shot & 61.31\% & 0.6928 & 0.6214 & 0.1699 & 0.1131 \\
      Gemini-2.5 Pro Zero-Shot~\cite{comanici2025gemini25pushingfrontier} & 60.28\% & 0.6856 & 0.6245 & 0.1702 & 0.1121 \\
      \hdashline
      Qwen2.5-VL-7B~\cite{Qwen2.5-VL} & 57.76\% & 0.6658 & 0.6110 & 0.1678 & 0.0733 \\
      Qwen2.5-VL-7B + GRPO & 58.97\% & 0.7020 & 0.6592 & 0.1741& 0.1003 \\
      Qwen2.5-VL-7B + SFT & 78.34\% & 0.8405 & 0.7768 & 0.4011& 0.3515 \\
      \rowcolor{gray!20}
      Ours & \textbf{80.10\%} & \textbf{0.8426} & \textbf{0.7785} & \textbf{0.4037}& \textbf{0.3530} \\
      \hdashline
      GLM-4.1V-9B~\cite{vteam2025glm45vglm41vthinkingversatilemultimodal} & 58.25\% & 0.6723 & 0.6189 & 0.1701 & 0.0966 \\
      GLM-4.1V-9B + GRPO & 60.15\% & 0.7182 & 0.6235 & 0.1833& 0.1226 \\
      GLM-4.1V-9B + SFT & 77.13\% & 0.8357 & 0.7734 & 0.3970 & 0.3592\\
      \rowcolor{gray!20}
      Ours & \textbf{79.26\%} & \textbf{0.8416} & \textbf{0.7811} & \textbf{0.4172}& \textbf{0.3757} \\
      \hdashline
      Ovis2.5-9B~\cite{lu2025ovis25technicalreport} & 56.94\% & 0.6801 & 0.6056 & 0.1678 & 0.1035 \\
      Ovis2.5-9B + GRPO & 69.67\% & 0.7264 & 0.6650 & 0.1901& 0.1563 \\
      Ovis2.5-9B + SFT & 78.85\% & 0.8287 & 0.7616 & 0.3589& 0.3286 \\
      \rowcolor{gray!20}
      Ours & \textbf{80.62\%} & \textbf{0.8392} & \textbf{0.7730} & \textbf{0.3865}& \textbf{0.3521} \\
      \bottomrule[1pt]
    \end{tabular}
  }
  \label{table:VLM}
  \vspace{-15pt}
\end{table}
\textbf{Human-Alignment Reasoning.} Tab.~\ref{table:VLM} reports quantitative results for the proposed Human-Alignment Reasoning Agent under different training strategies. Within each model family (Qwen2.5-VL, GLM-4.1V, Ovis2.5), our method consistently achieves the highest scores across all metrics, confirming that progressive alignment improves reasoning accuracy and adaptation to human preferences, effectively bridges distortion perception and linguistic reasoning within the \textit{perception-reasoning-action} loop.

Beyond validating the reasoning agent itself, these results also highlight the diagnostic value of our GenBlemish-27K dataset.
Across three large-scale backbones, performance consistently improves when trained or evaluated on our dataset, indicating its strong capability to reveal distortion-related reasoning gaps and to guide human-aligned adaptation.
Conversely, Zero-Shot settings, including advanced closed-source models such as GPT-5 and Gemini 2.5 Pro, exhibit limited generalization, underscoring the intrinsic difficulty of distortion-type reasoning and further validating the dataset’s discriminative and instructional effectiveness.

\subsection{Ablation Studies}




We conduct three ablation analyses across the Perception, Reasoning, and Action Agents to evaluate the contribution of each component within our framework. Specifically, we isolate (i) the lightweight attention mechanism and the KLD loss in the Perception Agent, (ii) progressive alignment strategies in the Reasoning Agent, and (iii) adaptive conditioning schemes in the Action Agent.


\textbf{Perception Agent Ablation.} Tab.~\ref{tab:Ablation_1} analyzes the effect of the lightweight attention mechanism and the KLD loss in the Perception Agent.
Removing attention (“w/o attn”) results in lower SIM and CC, indicating reduced global structural consistency.
Eliminating the KLD loss (“w/o KLD loss”) decreases NSS and AUC-Judd, suggesting less accurate fixation-level localization.
These two components are complementary: the attention module maintains coherent contextual structure, while the KLD term sharpens focus on human-attended regions.
Their joint optimization yields the best overall balance between local precision and global awareness, validating the perception agent’s contribution to context-aware saliency modeling.
\begin{table}[t]
  \centering
  \caption{Ablation study of the Context-Aware Perception Agent on attention and KLD loss components.}
  \renewcommand\arraystretch{1.2}  
  \resizebox{0.5\textwidth}{!}{  
    \begin{tabular}{lccccccc}  
      \toprule[1pt]
      \textbf{Method/Metric} & \textbf{AUC-Judd} $\uparrow$ & \textbf{ NSS}$\uparrow$ & \textbf{ CC} $\uparrow$& \textbf{ SIM}$\uparrow$ &\textbf{ KLD}$\downarrow$ \\
      \midrule
        Ours w/o attn\&KLD\_loss & \underline{0.9335} & 1.1957 & 0.5518 & \underline{0.3766} & 1.4436\\
        Ours w/o attn & \underline{0.9335} & \textbf{1.2153} & 0.5544 & 0.3731 & \underline{1.4412} \\
        Ours w/o KLD\_loss & 0.9313 & 1.1892 & \underline{0.5546} & 0.3525 & 1.5008\\
      \hdashline
      \rowcolor{gray!20} 
       Ours & \textbf{0.9336} & \underline{1.2087}  & \textbf{0.5568} & \textbf{0.3822} & \textbf{1.4313}\\
      \bottomrule[1pt]
    \end{tabular}
  }
  \label{tab:Ablation_1}
  \vspace{-15pt}
\end{table}

\textbf{Reasoning Agent Ablation.} Tab.~\ref{table:VLM} provides an ablation on reasoning alignment strategies.
Progressive training consistently outperforms single-stage SFT or GRPO-only configurations across all metrics. Notably, applying GRPO at early stages destabilizes response formatting and causes factual drift, whereas progressive alignment enhances both reasoning stability and human-aligned semantic grounding.

\textbf{Action Agent Ablation.} Tab.~\ref{tab:table1} compares different conditioning schemes for the Action Agent under both VLM-based and mask-based refinement settings.
Across all datasets, every tool in our tool library consistently achieves higher scores on all metrics, confirming robustness to diverse distortion types.
By dynamically selecting among multiple refinement backbones, the Action Agent ensures locally correction and global coherence, effectively closing the \textit{perception-reasoning-action} loop.




\section{Conclusions}

We propose Agentic Retoucher, a hierarchical, decision-driven framework that reformulates post-generation editing for T2I diffusion as a human-like \textit{perception-reasoning-action} loop. The perception agent localizes small-scale distortions, the reasoning agent performs human-aligned diagnosis, and the action agent plans localized inpainting guided by user intent. In addition, we introduce GenBlemish-27K for fine-grained supervision and evaluation. Extensive experiments demonstrate consistent improvements over state-of-the-art methods in perceptual quality, distortion localization, and human preference alignment, establishing a self-corrective and perceptually reliable T2I paradigm.

\section*{Acknowledgment}
This work is supported by National Natural Science Foundation of China (62571322, 62431015, 62271308), STCSM (24ZR1432000, 24511106902, 24511106900, 22DZ2229005), 111 plan (BP0719010), and State Key Laboratory of UHD Video and Audio Production and Presentation.


{
    \small
    \bibliographystyle{ieeenat_fullname}
    \bibliography{main}

@String(CVPR= {IEEE Conf. Comput. Vis. Pattern Recog.})

@String(ICCV= {Int. Conf. Comput. Vis.})

@String(NIPS= {Adv. Neural Inform. Process. Syst.})

@String(ICPR = {Int. Conf. Pattern Recog.})

@String(TIP  = {IEEE Trans. Image Process.})

@String(ICLR = {Int. Conf. Learn. Represent.})

@String(CVPR  = {CVPR})

@String(ICCV  = {ICCV})

@String(NIPS  = {NeurIPS})

@String(ICPR  = {ICPR})

@String(TIP   = {IEEE TIP})

@String(ICLR  = {ICLR})

@article{Wang2024HADM,
  title={Detecting Human Artifacts from Text-to-Image Models},
  author={Wang, Kaihong and Zhang, Lingzhi and Zhang, Jianming},
  journal={arXiv preprint arXiv:2411.13842},
  year={2024}
}

@InProceedings{Wang_2025_CVPR,
    author    = {Wang, Zeqing and Ma, Qingyang and Wan, Wentao and Li, Haojie and Wang, Keze and Tian, Yonghong},
    title     = {Is this Generated Person Existed in Real-world? Fine-grained Detecting and Calibrating Abnormal Human-body},
    booktitle = {Proceedings of the IEEE/CVF CVPR},
    month     = {June},
    year      = {2025},
    pages     = {21226-21237}
}

@InProceedings{Liang_2024_CVPR,
    author    = {Liang, Youwei and He, Junfeng and Li, Gang and Li, Peizhao and Klimovskiy, Arseniy and Carolan, Nicholas et al.},
    title     = {Rich Human Feedback for Text-to-Image Generation},
    booktitle = {Proceedings of the IEEE/CVF CVPR},
    month     = {June},
    year      = {2024},
    pages     = {19401-19411}
}

@article{cao2024synartifact,
  title={SynArtifact: Classifying and Alleviating Artifacts in Synthetic Images via Vision-Language Model},
  author={Cao, Bin and Yuan, Jianhao and Liu, Yexin and Li, Jian and Sun, Shuyang and Liu, Jing and Zhao, Bo},
  journal={arXiv preprint arXiv:2402.18068},
  year={2024}
}

@misc{zhang2024abench,
      title={A-Bench: Are LMMs Masters at Evaluating AI-generated Images?}, 
      author={Zicheng Zhang and Haoning Wu and Chunyi Li and Yingjie Zhou and Wei Sun and Xiongkuo Min and Zijian Chen and Xiaohong Liu and Weisi Lin and Guangtao Zhai},
      year={2024},
      eprint={2406.03070},
      archivePrefix={arXiv},
      primaryClass={cs.CV}
}

@misc{han2025ntire2025challengetext,
      title={NTIRE 2025 challenge on Text to Image Generation Model Quality Assessment}, 
      year={2025},
      eprint={2505.16314},
      archivePrefix={arXiv},
      primaryClass={cs.CV},
      url={https://arxiv.org/abs/2505.16314}, 
}

@ARTICLE{8315047,
  author={Bylinskii, Zoya and Judd, Tilke and Oliva, Aude and Torralba, Antonio and Durand, Frédo},
  journal={IEEE TPAMI}, 
  title={What Do Different Evaluation Metrics Tell Us About Saliency Models?}, 
  year={2019},
  volume={41},
  number={3},
  pages={740-757},
  doi={10.1109/TPAMI.2018.2815601}}

@misc{wu2025qwenimagetechnicalreport,
      title={Qwen-Image Technical Report}, 
      year={2025},
      eprint={2508.02324},
      archivePrefix={arXiv},
      primaryClass={cs.CV},
      url={https://arxiv.org/abs/2508.02324}, 
}

@misc{flux2024,
    author={Black Forest Labs},
    title={FLUX},
    year={2024},
    howpublished={},
}

@article{podell2023sdxl,
  title={Sdxl: Improving latent diffusion models for high-resolution image synthesis},
  author={Podell, Dustin and English, Zion and Lacey, Kyle and Blattmann, Andreas and Dockhorn, Tim and M{\"u}ller, Jonas and Penna, Joe and Rombach, Robin},
  journal={arXiv preprint arXiv:2307.01952},
  year={2023}
}

@InProceedings{wu2024megafusion,
        author    = {Wu, Haoning and Shen, Shaocheng and Hu, Qiang and Zhang, Xiaoyun and Zhang, Ya and Wang, Yanfeng},
        title     = {MegaFusion: Extend Diffusion Models towards Higher-resolution Image Generation without Further Tuning},
        booktitle = {WACV},
        year      = {2025},
  }

@article{saharia2022photorealistic,
  title={Photorealistic text-to-image diffusion models with deep language understanding},
  author={Saharia, Chitwan and Chan, William and Saxena, Saurabh and Li, Lala and Whang, Jay and Denton, Emily L and Ghasemipour, Kamyar and Gontijo Lopes, Raphael and Karagol Ayan, Burcu and Salimans, Tim and others},
  journal={NIPS},
  volume={35},
  pages={36479--36494},
  year={2022}
}

@inproceedings{rombach2022high,
  title={High-resolution image synthesis with latent diffusion models},
  author={Rombach, Robin and Blattmann, Andreas and Lorenz, Dominik and Esser, Patrick and Ommer, Bj{\"o}rn},
  booktitle={Proceedings of the IEEE/CVF CVPR},
  pages={10684--10695},
  year={2022}
}

@inproceedings{brooks2023instructpix2pix,
  title={Instructpix2pix: Learning to follow image editing instructions},
  author={Brooks, Tim and Holynski, Aleksander and Efros, Alexei A},
  booktitle={CVPR},
  pages={18392--18402},
  year={2023}
}

@inproceedings{kawar2023imagic,
  title={Imagic: Text-based real image editing with diffusion models},
  author={Kawar, Bahjat and Zada, Shiran and Lang, Oran and Tov, Omer and Chang, Huiwen and Dekel, Tali and Mosseri, Inbar and Irani, Michal},
  booktitle={Proceedings of the IEEE/CVF CVPR},
  pages={6007--6017},
  year={2023}
}

@article{deng2025bagel,
  title   = {Emerging Properties in Unified Multimodal Pretraining},
  author  = {Deng, Chaorui and Zhu, Deyao and Li, Kunchang and Gou, Chenhui and Li, Feng and Wang, Zeyu and Zhong, Shu and Yu, Weihao and Nie, Xiaonan and Song, Ziang and Shi, Guang and Fan, Haoqi},
  journal = {arXiv preprint arXiv:2505.14683},
  year    = {2025}
}

@article{liu2025step1x-edit,
      title={Step1X-Edit: A Practical Framework for General Image Editing}, 
      author={Shiyu Liu and Yucheng Han et al.},
      journal={arXiv preprint arXiv:2504.17761},
      year={2025}
}

@article{wan2025,
      title={Wan: Open and Advanced Large-Scale Video Generative Models}, 
      author={Team Wan and Ang Wang and Baole Ai et al.},
      journal = {arXiv preprint arXiv:2503.20314},
      year={2025}
}

@article{opensora,
  title={Open-sora: Democratizing efficient video production for all},
  author={Zheng, Zangwei and Peng, Xiangyu and Yang, Tianji and Shen, Chenhui and Li, Shenggui and Liu, Hongxin and Zhou, Yukun and Li, Tianyi and You, Yang},
  journal={arXiv preprint arXiv:2412.20404},
  year={2024}
}

@misc{chen2025sanavideoefficientvideogeneration,
      title={SANA-Video: Efficient Video Generation with Block Linear Diffusion Transformer}, 
      author={Junsong Chen and Yuyang Zhao and Jincheng Yu and Ruihang Chu and Junyu Chen and Shuai Yang and Xianbang Wang and Yicheng Pan and Daquan Zhou and Huan Ling and Haozhe Liu and Hongwei Yi and Hao Zhang and Muyang Li and Yukang Chen and Han Cai and Sanja Fidler and Ping Luo and Song Han and Enze Xie},
      year={2025},
      eprint={2509.24695},
      archivePrefix={arXiv},
      primaryClass={cs.CV},
      url={https://arxiv.org/abs/2509.24695}, 
}

@misc{wang2025promptenhancersimpleapproachenhance,
      title={PromptEnhancer: A Simple Approach to Enhance Text-to-Image Models via Chain-of-Thought Prompt Rewriting}, 
      author={Linqing Wang and Ximing Xing et al.},
      year={2025},
      eprint={2509.04545},
      archivePrefix={arXiv},
      primaryClass={cs.CV},
      url={https://arxiv.org/abs/2509.04545}, 
}

@inproceedings{black2023ddpo,
      title={Training Diffusion Models with Reinforcement Learning},
      author={Kevin Black and Michael Janner and Yilun Du and Ilya Kostrikov and Sergey Levine},
      year={2023},
      eprint={2305.13301},
      archivePrefix={arXiv},
      primaryClass={cs.LG}
}

@misc{arkhipkin2023kandinsky,
      title={Kandinsky 3.0 Technical Report}, 
      author={Vladimir Arkhipkin and Andrei Filatov and Viacheslav Vasilev and Anastasia Maltseva and Said Azizov and Igor Pavlov and Julia Agafonova and Andrey Kuznetsov and Denis Dimitrov},
      year={2023},
      eprint={2312.03511},
      archivePrefix={arXiv},
      primaryClass={cs.CV}
}

@misc{yang2025qwen3technicalreport,
      title={Qwen3 Technical Report}, 
      author={An Yang and Anfeng Li et al.},
      year={2025},
      eprint={2505.09388},
      archivePrefix={arXiv},
      primaryClass={cs.CL},
      url={https://arxiv.org/abs/2505.09388}, 
}

@article{Qwen2.5-VL,
  title={Qwen2.5-VL Technical Report},
  author={Bai, Shuai and Chen et al.},
  journal={arXiv preprint arXiv:2502.13923},
  year={2025}
}

@article{2020An,
  title={An Image is Worth 16x16 Words: Transformers for Image Recognition at Scale},
  author={ Dosovitskiy, Alexey  and  Beyer, Lucas  and  Kolesnikov, Alexander  and  Weissenborn, Dirk  and  Houlsby, Neil },
  year={2020},
}

@misc{roberts2022scalingmodelsdatatextttt5x,
      title={Scaling Up Models and Data with $\texttt{t5x}$ and $\texttt{seqio}$}, 
      author={Adam Roberts and Hyung Won Chung et al.},
      year={2022},
      eprint={2203.17189},
      archivePrefix={arXiv},
      primaryClass={cs.LG},
      url={https://arxiv.org/abs/2203.17189}, 
}

@inproceedings{hu2022lowrank,
  author = {Hu, Edward J. and Shen, Yelong and Wallis, Phillip and Allen-Zhu, Zeyuan and Li, Yuanzhi and Wang, Shean and Wang, Lu and Chen, Weizhu},
  booktitle = {ICLR},
  ee = {https://openreview.net/forum?id=nZeVKeeFYf9},
  publisher = {OpenReview.net},
  title = {LoRA: Low-Rank Adaptation of Large Language Models.},
  url = {http://dblp.uni-trier.de/db/conf/iclr/iclr2022.html#HuSWALWWC22},
  year = 2022
}

@misc{deepseekai2025deepseekr1incentivizingreasoningcapability,
      title={DeepSeek-R1: Incentivizing Reasoning Capability in LLMs via Reinforcement Learning}, 
      author={DeepSeek-AI and Daya Guo and Dejian Yang et al.},
      year={2025},
      eprint={2501.12948},
      archivePrefix={arXiv},
      primaryClass={cs.CL},
      url={https://arxiv.org/abs/2501.12948}, 
}

@inproceedings{lin-2004-rouge,
    title = "{ROUGE}: A Package for Automatic Evaluation of Summaries",
    author = "Lin, Chin-Yew",
    booktitle = "Text Summarization Branches Out",
    month = jul,
    year = "2004",
    address = "Barcelona, Spain",
    publisher = "Association for Computational Linguistics",
    url = "https://aclanthology.org/W04-1013/",
    pages = "74--81"
}

@inproceedings{banerjee-lavie-2005-meteor,
    title = "{METEOR}: An Automatic Metric for {MT} Evaluation with Improved Correlation with Human Judgments",
    author = "Banerjee, Satanjeev  and
      Lavie, Alon",
    editor = "Goldstein, Jade  and
      Lavie, Alon  and
      Lin, Chin-Yew  and
      Voss, Clare",
    booktitle = "Proceedings of the {ACL} Workshop on Intrinsic and Extrinsic Evaluation Measures for Machine Translation and/or Summarization",
    month = jun,
    year = "2005",
    url = "https://aclanthology.org/W05-0909/",
    pages = "65--72"
}

@misc{mikolov2013efficientestimationwordrepresentations,
      title={Efficient Estimation of Word Representations in Vector Space}, 
      author={Tomas Mikolov and Kai Chen and Greg Corrado and Jeffrey Dean},
      year={2013},
      eprint={1301.3781},
      archivePrefix={arXiv},
      primaryClass={cs.CL},
      url={https://arxiv.org/abs/1301.3781}, 
}

@inproceedings{gao2021simcse,
   title={{SimCSE}: Simple Contrastive Learning of Sentence Embeddings},
   author={Gao, Tianyu and Yao, Xingcheng and Chen, Danqi},
   booktitle={EMNLP},
   year={2021}
}

@inproceedings{NIPS2005_0738069b,
 author = {Bruce, Neil and Tsotsos, John},
 booktitle = {NIPS},
 editor = {Y. Weiss and B. Sch\"{o}lkopf and J. Platt},
 pages = {},
 publisher = {MIT Press},
 title = {Saliency Based on Information Maximization},
 url = {https://proceedings.neurips.cc/paper_files/paper/2005/file/0738069b244a1c43c83112b735140a16-Paper.pdf},
 volume = {18},
 year = {2005}
}

@inproceedings{NIPS2006_4db0f8b0,
 author = {Harel, Jonathan and Koch, Christof and Perona, Pietro},
 booktitle = {NIPS},
 editor = {B. Sch\"{o}lkopf and J. Platt and T. Hoffman},
 pages = {},
 publisher = {MIT Press},
 title = {Graph-Based Visual Saliency},
 url = {https://proceedings.neurips.cc/paper_files/paper/2006/file/4db0f8b0fc895da263fd77fc8aecabe4-Paper.pdf},
 volume = {19},
 year = {2006}
}

@INPROCEEDINGS{4270292,
  author={Hou, Xiaodi and Zhang, Liqing},
  booktitle={2007 IEEE CVPR}, 
  title={Saliency Detection: A Spectral Residual Approach}, 
  year={2007},
  volume={},
  number={},
  pages={1-8},
  doi={10.1109/CVPR.2007.383267}}

@inproceedings{NIPS2007_708f3cf8,
 author = {Cerf, Moran and Harel, Jonathan and Einhaeuser, Wolfgang and Koch, Christof},
 booktitle = {NIPS},
 editor = {J. Platt and D. Koller and Y. Singer and S. Roweis},
 pages = {},
 publisher = {Curran Associates, Inc.},
 title = {Predicting human gaze using low-level saliency combined with face detection},
 url = {https://proceedings.neurips.cc/paper_files/paper/2007/file/708f3cf8100d5e71834b1db77dfa15d6-Paper.pdf},
 volume = {20},
 year = {2007}
}

@inproceedings{zhao2011visual,
  title={Visual saliency detection by spatially weighted dissimilarity},
  author={Zhao, Qing and Cai, Jianfei},
  booktitle={2011 IEEE CVPR},
  pages={1241--1248},
  year={2011},
  organization={IEEE}
}

@INPROCEEDINGS{5539929,
  author={Goferman, Stas and Zelnik-Manor, Lihi and Tal, Ayellet},
  booktitle={2010 IEEE Computer Society Conference on Computer Vision and Pattern Recognition}, 
  title={Context-aware saliency detection}, 
  year={2010},
  volume={},
  number={},
  pages={2376-2383},
  doi={10.1109/CVPR.2010.5539929}}

@InProceedings{Huang_2015_ICCV,
author = {Huang, Xun and Shen, Chengyao and Boix, Xavier and Zhao, Qi},
title = {SALICON: Reducing the Semantic Gap in Saliency Prediction by Adapting Deep Neural Networks},
booktitle = {ICCV},
month = {December},
year = {2015}
}

@article{Lou2022TranSalNetTP,
  title={TranSalNet: Towards perceptually relevant visual saliency prediction},
  author={Jovo Lou and Lin Ma and Kai-Xin Hu and Huazhe Yang and Wen-Yan Lin},
  journal={Neurocomputing},
  year={2022},
  volume={507},
  pages={250-264}
}

@article{Che_2020,
   title={How is Gaze Influenced by Image Transformations? Dataset and Model},
   volume={29},
   ISSN={1941-0042},
   url={http://dx.doi.org/10.1109/TIP.2019.2945857},
   DOI={10.1109/tip.2019.2945857},
   journal={IEEE TIP},
   publisher={Institute of Electrical and Electronics Engineers (IEEE)},
   author={Che, Zhaohui and Borji, Ali and Zhai, Guangtao and Min, Xiongkuo and Guo, Guodong and Le Callet, Patrick},
   year={2020},
   pages={2287–2300} }

@article{cornia2018predicting,
  author = {Cornia, Marcella and Baraldi, Lorenzo and Serra, Giuseppe and Cucchiara, Rita},
  title = {{Predicting Human Eye Fixations via an LSTM-based Saliency Attentive Model}},
  journal = {IEEE Transactions on Image Processing},
  volume={27},
  number={10},
  pages={5142--5154},
  year = {2018}
}

@inproceedings{mlnet2016,
  author = {Cornia, Marcella and Baraldi, Lorenzo and Serra, Giuseppe and Cucchiara, Rita},
  title = {{A Deep Multi-Level Network for Saliency Prediction}},
  booktitle = {ICPR},
  year = {2016}
}

@misc{vteam2025glm45vglm41vthinkingversatilemultimodal,
      title={GLM-4.5V and GLM-4.1V-Thinking: Towards Versatile Multimodal Reasoning with Scalable Reinforcement Learning}, 
      author={V Team and Wenyi Hong et al.},
      year={2025},
      eprint={2507.01006},
      archivePrefix={arXiv},
      primaryClass={cs.CV},
      url={https://arxiv.org/abs/2507.01006}, 
}

@misc{wang2025internvl35advancingopensourcemultimodal,
      title={InternVL3.5: Advancing Open-Source Multimodal Models in Versatility, Reasoning, and Efficiency}, 
      author={Weiyun Wang and Zhangwei Gao et al.},
      year={2025},
      eprint={2508.18265},
      archivePrefix={arXiv},
      primaryClass={cs.CV},
      url={https://arxiv.org/abs/2508.18265}, 
}

@misc{lu2025ovis25technicalreport,
      title={Ovis2.5 Technical Report}, 
      author={Shiyin Lu and Yang Li et.al},
      year={2025},
      eprint={2508.11737},
      archivePrefix={arXiv},
      primaryClass={cs.CV},
      url={https://arxiv.org/abs/2508.11737}, 
}

@misc{comanici2025gemini25pushingfrontier,
      title={Gemini 2.5: Pushing the Frontier with Advanced Reasoning, Multimodality, Long Context, and Next Generation Agentic Capabilities}, 
      author={Gheorghe Comanici and Eric Bieber and Mike Schaekermann et al.},
      year={2025},
      eprint={2507.06261},
      archivePrefix={arXiv},
      primaryClass={cs.CL},
      url={https://arxiv.org/abs/2507.06261}, 
}

@inproceedings{NIPS2017_8a1d6947,
 author = {Heusel, Martin and Ramsauer, Hubert and Unterthiner, Thomas and Nessler, Bernhard and Hochreiter, Sepp},
 booktitle = {NIPS},
 editor = {I. Guyon and U. Von Luxburg and S. Bengio and H. Wallach and R. Fergus and S. Vishwanathan and R. Garnett},
 pages = {},
 publisher = {Curran Associates, Inc.},
 title = {GANs Trained by a Two Time-Scale Update Rule Converge to a Local Nash Equilibrium},
 url = {https://proceedings.neurips.cc/paper_files/paper/2017/file/8a1d694707eb0fefe65871369074926d-Paper.pdf},
 volume = {30},
 year = {2017}
}

@article{jarvisart2025,
title={JarvisArt: Liberating Human Artistic Creativity via an Intelligent Photo Retouching Agent}, 
      author={Yunlong Lin and Zixu Lin and Kunjie Lin and Jinbin Bai and Panwang Pan and Chenxin Li and Haoyu Chen and Zhongdao Wang and Xinghao Ding and Wenbo Li and Shuicheng Yan},
      year={2025},
      journal={arXiv preprint arXiv:2506.17612}
}

@misc{liu2025moavrmixtureofagentsallinonevideo,
      title={MoA-VR: A Mixture-of-Agents System Towards All-in-One Video Restoration}, 
      author={Lu Liu and Chunlei Cai and Shaocheng Shen and Jianfeng Liang and Weimin Ouyang and Tianxiao Ye and Jian Mao and Huiyu Duan and Jiangchao Yao and Xiaoyun Zhang and Qiang Hu and Guangtao Zhai},
      year={2025},
      eprint={2510.08508},
      archivePrefix={arXiv},
      primaryClass={cs.CV},
      url={https://arxiv.org/abs/2510.08508}, 
}

@inproceedings{agenticir,
      title={An Intelligent Agentic System for Complex Image Restoration Problems},
      author={Kaiwen Zhu and Jinjin Gu and Zhiyuan You and Yu Qiao and Chao Dong},
      booktitle={The Thirteenth ICLR},
      year={2025},
}

@misc{izadi2025finegrainedalignmentnoiserefinement,
      title={Fine-Grained Alignment and Noise Refinement for Compositional Text-to-Image Generation}, 
      author={Amir Mohammad Izadi and Seyed Mohammad Hadi Hosseini and Soroush Vafaie Tabar and Ali Abdollahi and Armin Saghafian and Mahdieh Soleymani Baghshah},
      year={2025},
      eprint={2503.06506},
      archivePrefix={arXiv},
      primaryClass={cs.CV},
      url={https://arxiv.org/abs/2503.06506}, 
}

@article{Le2025FromRT,
  title={From Reflection to Perfection: Scaling Inference-Time Optimization for Text-to-Image Diffusion Models via Reflection Tuning},
  author={Yixuan Le and Yujun Shen and Bolei Zhou},
  journal={arXiv preprint arXiv:2504.16080},
  year={2025}
}

@article{Cubuk2019AutoAugmentLA,
  title={AutoAugment: Learning Augmentation Policies from Data},
  author={Ekin D. Cubuk and Barret Zoph and Dandelion Man'e and Vijay Vasudevan and Quoc V. Le},
  journal={2019 IEEE/CVF CVPR},
  year={2019},
  pages={113-123}
}

@inproceedings{wang2021tent,
  title={Tent: Fully Test-Time Adaptation by Entropy Minimization},
  author={Wang, Dequan and Shelhamer, Evan and Liu, Shaoteng and Olshausen, Bruno and Darrell, Trevor},
  booktitle={ICLR},
  year={2021},
  url={https://openreview.net/forum?id=uXl3bZLkr3c}
}

@ARTICLE{538972,
  author={Hutchinson, S. and Hager, G.D. and Corke, P.I.},
  journal={IEEE Transactions on Robotics and Automation}, 
  title={A tutorial on visual servo control}, 
  year={1996},
  volume={12},
  number={5},
  pages={651-670},
  doi={10.1109/70.538972}}

@article{RotherKB04,
  author    = {Carsten Rother and
               Vladimir Kolmogorov and
               Andrew Blake},
  title     = {``GrabCut'': interactive foreground extraction using iterated graph cuts},
  journal   = {ACM Trans. Graph.},
  volume    = {23},
  number    = {3},
  pages     = {309--314},
  year      = {2004},
  url       = {https://doi.org/10.1145/1186562.1015720},
  doi       = {10.1145/1186562.1015720},
  publisher = {ACM},
  acmid     = {1015720},
  address   = {New York, NY, USA},
}

@article{zhu2023ghost,
  title={Ghost in the Minecraft: Generally Capable Agents for Open-World Environments via Large Language Models with Text-based Knowledge and Memory},
  author={Zhu, Xizhou and Chen, Yuntao and Tian, Hao and Tao, Chenxin and Su, Weijie and Yang, Chenyu and Huang, Gao and Li, Bin and Lu, Lewei and Wang, Xiaogang and Qiao, Yu and Zhang, Zhaoxiang and Dai, Jifeng},
  journal={arXiv preprint arXiv:2305.17144},
  year={2023}
}

@article{li2023generalist,
    title={Towards Generalist Robot Policies: What Matters in Building Vision-Language-Action Models},
    author={Li, Xinghang and Li, Peiyan and Liu, Minghuan and Wang, Dong and Liu, Jirong and Kang, Bingyi and Ma, Xiao and Kong, Tao and Zhang, Hanbo and Liu, Huaping},
    journal={arXiv preprint arXiv:2412.14058},
    year={2024}
}

@misc{chen2024restoreagent,
    title={RestoreAgent: Autonomous Image Restoration Agent via Multimodal Large Language Models},
    author={Haoyu Chen and Wenbo Li and Jinjin Gu and Jingjing Ren and Sixiang Chen and Tian Ye and Renjing Pei and Kaiwen Zhou and Fenglong Song and Lei Zhu},
    year={2024},
    eprint={2407.18035},
    archivePrefix={arXiv},
    primaryClass={cs.CV}
}

@misc{GPT4,
  author       = {OpenAI Team},
  title        = {ChatGPT-4o},
  year         = {2024},
  note         = {Accessed: 2025-03-08}
}

@misc{qian2025explainablepartialaigcimagequality,
      title={Towards Explainable Partial-AIGC Image Quality Assessment}, 
      author={Jiaying Qian and Ziheng Jia and Zicheng Zhang and Zeyu Zhang and Guangtao Zhai and Xiongkuo Min},
      year={2025},
      eprint={2504.09291},
      archivePrefix={arXiv},
      primaryClass={cs.CV},
      url={https://arxiv.org/abs/2504.09291}, 
}

@misc{chang2025oneigbenchomnidimensionalnuancedevaluation,
      title={OneIG-Bench: Omni-dimensional Nuanced Evaluation for Image Generation}, 
      author={Jingjing Chang and Yixiao Fang and Peng Xing and Shuhan Wu and Wei Cheng and Rui Wang and Xianfang Zeng and Gang Yu and Hai-Bao Chen},
      year={2025},
      eprint={2506.07977},
      archivePrefix={arXiv},
      primaryClass={cs.CV},
      url={https://arxiv.org/abs/2506.07977}, 
}

@misc{wang2025cigeval,
      title={A Unified Agentic Framework for Evaluating Conditional Image Generation}, 
      author={Jifang Wang and Xue Yang and Longyue Wang and Zhenran Xu and Yiyu Wang and Yaowei Wang and Weihua Luo and Kaifu Zhang and Baotian Hu and Min Zhang},
      year={2025},
      eprint={2504.07046},
      archivePrefix={arXiv},
      primaryClass={cs.CV},
      url={https://arxiv.org/abs/2504.07046}, 
}

@article{hertz2022prompt,
  title={Prompt-to-prompt image editing with cross attention control},
  author={Hertz, Amir and Mokady, Ron and Tenenbaum, Jay and Aberman, Kfir and Pritch, Yael and Cohen-Or, Daniel},
  booktitle={arXiv preprint arXiv:2208.01626},
  year={2022}
}

@misc{openai2024gpt4ocard,
      title={GPT-4o System Card}, 
      author={OpenAI and Aaron Hurst et al.},
      year={2024},
      eprint={2410.21276},
      archivePrefix={arXiv},
      primaryClass={cs.CL},
      url={https://arxiv.org/abs/2410.21276}, 
}

@misc{li2024llavaonevisioneasyvisualtask,
      title={LLaVA-OneVision: Easy Visual Task Transfer}, 
      author={Bo Li and Yuanhan Zhang and Dong Guo and Renrui Zhang and Feng Li and Hao Zhang and Kaichen Zhang and Peiyuan Zhang and Yanwei Li and Ziwei Liu and Chunyuan Li},
      year={2024},
      eprint={2408.03326},
      archivePrefix={arXiv},
      primaryClass={cs.CV},
      url={https://arxiv.org/abs/2408.03326}, 
}

@article{wu2025reprompt,
  title={RePrompt: Reasoning-Augmented Reprompting for Text-to-Image Generation via Reinforcement Learning},
  author={Wu, Mingrui and Wang, Lu and Zhao, Pu and Yang, Fangkai and Zhang, Jianjin and Liu, Jianfeng and Zhan, Yuefeng and Han, Weihao and Sun, Hao and Ji, Jiayi and others},
  journal={arXiv preprint arXiv:2505.17540},
  year={2025}
}

@misc{kaduri2024_vision_of_vlms,
                title={What's in the Image? A Deep-Dive into the Vision of Vision Language Models}, 
                author={Omri Kaduri and Shai Bagon and Tali Dekel},
                year={2024},
                eprint={2411.17491},
                archivePrefix={arXiv},
                primaryClass={cs.CV},
                url={https://arxiv.org/abs/2411.17491}, 
}

@InProceedings{Qu_2025_CVPR,
    author    = {Qu, Liao and Zhang, Huichao and Liu, Yiheng and Wang, Xu and Jiang, Yi and Gao, Yiming and Ye, Hu and Du, Daniel K. and Yuan, Zehuan and Wu, Xinglong},
    title     = {TokenFlow: Unified Image Tokenizer for Multimodal Understanding and Generation},
    booktitle = {Proceedings of the Computer Vision and Pattern Recognition Conference (CVPR)},
    month     = {June},
    year      = {2025},
    pages     = {2545-2555}
}

@InProceedings{Heo_2025_CVPR,
    author    = {Heo, Miran and Chen, Min-Hung and Huang, De-An and Liu, Sifei and Radhakrishnan, Subhashree and Kim, Seon Joo and Wang, Yu-Chiang Frank and Hachiuma, Ryo},
    title     = {Omni-RGPT: Unifying Image and Video Region-level Understanding via Token Marks},
    booktitle = {Proceedings of the Computer Vision and Pattern Recognition Conference (CVPR)},
    month     = {June},
    year      = {2025},
    pages     = {3919-3930}
}

@inproceedings{Marsili_2025_CVPR,
    author    = {Marsili, Damiano and Agrawal, Rohun and Yue, Yisong and Gkioxari, Georgia},
    title     = {Visual Agentic AI for Spatial Reasoning with a Dynamic API},
    booktitle = {Proceedings of the Computer Vision and Pattern Recognition Conference (CVPR)},
    month     = {June},
    year      = {2025},
    pages     = {19446-19455}
}

@article{fang2025one,
  title={One-Step Diffusion Transformer for Controllable Real-World Image Super-Resolution},
  author={Fang, Yushun and Chen, Yuxiang and Yin, Shibo and Hu, Qiang and Yao, Jiangchao and Zhang, Ya and Zhang, Xiaoyun and Wang, Yanfeng},
  journal={arXiv preprint arXiv:2511.17138},
  year={2025}
}

@article{zhang2025lato,
  title={LaTo: Landmark-tokenized Diffusion Transformer for Fine-grained Human Face Editing},
  author={Zhang, Zhenghao and Zhang, Ziying and Liao, Junchao and Meng, Xiangyu and Hu, Qiang and Zhu, Siyu and Zhang, Xiaoyun and Qin, Long and Wang, Weizhi},
  journal={arXiv preprint arXiv:2509.25731},
  year={2025}
}

@article{liang2026d,
  title = {{D}$^2$-VR: Degradation-Robust and Distilled Video Restoration with Synergistic Optimization Strategy},
  author = {Liang, Jianfeng and Shen, Shaocheng and Xu, Botao and Hu, Qiang and Zhang, Xiaoyun},
  journal = {arXiv preprint arXiv:2602.08395},
  year = {2026}
}

@inproceedings{fang2025robust,
  title={Robust ID-Specific Face Restoration via Alignment Learning},
  author={Fang, Yushun and Liu, Lu and Gao, Xiang and Hu, Qiang and Cao, Ning and Cui, Jianghe and Chen, Gang and Zhang, Xiaoyun},
  booktitle={Chinese Conference on Pattern Recognition and Computer Vision (PRCV)},
  pages={122--137},
  year={2025},
  organization={Springer}
}

@inproceedings{duan2025finevq,
  title={Finevq: Fine-grained user generated content video quality assessment},
  author={Duan, Huiyu and Hu, Qiang and Wang, Jiarui and Yang, Liu and Xu, Zitong and Liu, Lu and Min, Xiongkuo and Cai, Chunlei and Ye, Tianxiao and Zhang, Xiaoyun and others},
  booktitle={Proceedings of the Computer Vision and Pattern Recognition Conference},
  pages={3206--3217},
  year={2025}
}

@inproceedings{liu2025f,
  title={F-bench: Rethinking human preference evaluation metrics for benchmarking face generation, customization, and restoration},
  author={Liu, Lu and Duan, Huiyu and Hu, Qiang and Yang, Liu and Cai, Chunlei and Ye, Tianxiao and Liu, Huayu and Zhang, Xiaoyun and Zhai, Guangtao},
  booktitle={Proceedings of the IEEE/CVF International Conference on Computer Vision},
  pages={10982--10994},
  year={2025}
}
}

\end{document}